\documentclass[conference]{IEEEtran}
\IEEEoverridecommandlockouts
\usepackage{cite}
\usepackage{amsmath,amssymb,amsfonts}
\usepackage{algorithmic}
\usepackage{graphicx}
\usepackage{textcomp}
\usepackage{xcolor}
\usepackage{booktabs}
\usepackage{booktabs}
\usepackage[flushleft]{threeparttable}
\usepackage{multirow}
\def\BibTeX{{\rm B\kern-.05em{\sc i\kern-.025em b}\kern-.08em
    T\kern-.1667em\lower.7ex\hbox{E}\kern-.125emX}}
\begin{document}

\title{Enhancing Embedding Performance through Large Language Model-based Text Enrichment and Rewriting\\

}

\author{\IEEEauthorblockN{ Nicholas Harris}
\IEEEauthorblockA{\textit{} 
\textit{Arizona State University}\\
Tempe, Arizona\\
 nick@myautobio.com}

\and

\IEEEauthorblockN{
Anand Butani}
\IEEEauthorblockA{\textit{MyAutoBio Inc.} \\
\textit{Scottsdale, Arizona}\\
anand@myautobio.com}

\and

\IEEEauthorblockN{
Syed Hashmy}
\IEEEauthorblockA{\textit{} 
\textit{Arizona State University}\\
Tempe, Arizona \\
shashmy@asu.edu}

}

\maketitle

\begin{abstract}
Embedding models are crucial for various natural language processing tasks but can be limited by factors such as limited vocabulary, lack of context, and grammatical errors. This paper proposes a novel approach to improve embedding performance by leveraging large language models (LLMs) to enrich and rewrite input text before the embedding process. By utilizing ChatGPT $3.5$ to provide additional context, correct inaccuracies, and incorporate metadata, the proposed method aims to enhance the utility and accuracy of embedding models. The effectiveness of this approach is evaluated on three datasets: Banking77Classification, TwitterSemEval $2015$, and Amazon Counter-factual Classification. Results demonstrate significant improvements over the baseline model on the TwitterSemEval $2015$ dataset, with the best-performing prompt achieving a score of 85.34 compared to the previous best of $81.52$ on the Massive Text Embedding Benchmark (MTEB) Leaderboard. However, performance on the other two datasets was less impressive, highlighting the importance of considering domain-specific characteristics. The findings suggest that LLM-based text enrichment has shown promising results to improve embedding performance, particularly in certain domains. Hence, numerous limitations in the process of embedding can be avoided.

\end{abstract}


\section{Introduction} 
Text embeddings are widely adopted in the field of Natural Language Processing (NLP) that refer to vectorized representation of natural language. An embedding is a representation of words in a low-dimensional continuous vector space. They encapsulate the semantic content of the text \cite{pittaras2021text}. These embeddings find extensive applications across a spectrum of natural language processing (NLP) endeavors including information retrieval (IR), question answering, assessing semantic textual similarity, mining bitexts, recommending items, etc. The researchers are making continuous efforts to improve the accuracy and reduce the training steps \cite{wang2023improving}.

Furthermore, an efficient technique for creating high-quality text embeddings using synthetic data and minimal training, avoiding complex pipelines and extensive labeled datasets, and achieving top results on key benchmarks when mixed with labeled data \cite{wang2023improving}.

There were early approaches like word2vec \cite{mikolov2013efficient} and GloVe \cite{pennington2014glove} to more advanced models such as FastText \cite{bojanowski2017enriching} and BERT \cite{devlin2018bert}. It discusses the strengths and limitations of each model and their impact on various natural language processing (NLP) tasks.

Various techniques have been proposed to improve the performance of embedding models, such as fine-tuning on domain-specific data \cite{howard2018universal}, using ensemble methods, and incorporating external knowledge sources \cite{zhang2019ernie}. Large language models have been successfully applied to a wide range of NLP tasks, such as text generation \cite{radford2019language}, question answering \cite{raffel2020exploring}, and sentiment analysis \cite{brown2020language}. Several studies have explored the use of text enrichment and rewriting techniques to improve the quality and informativeness of text data. For example, a method for contextual augmentation of text data using a bidirectional language model is being proposed \cite{kobayashi2018contextual}, while a retrieval-augmented generation approach for improving the factual accuracy of generated text was also introduced \cite{guu2020retrieval}. 

Recent research has explored the use of LLMs for text compression to reduce computational costs in Retrieval-Augmented Generation (RAG) systems and large LLMs. For instance, RECOMP proposes compressing retrieved documents into summaries before integrating them with language models, aiming to reduce computational costs and help LMs identify relevant information more efficiently \cite{xu2023recomp}. Similarly, TCRA-LLM introduces a token compression scheme for retrieval-augmented LLMs, employing summarization and semantic compression techniques to reduce inference costs \cite{liu2023tcra}.
Context Tuning for RAG addresses the limitation of RAG's tool retrieval step by employing a smart context retrieval system to fetch relevant information, improving the efficiency and effectiveness of the generation process \cite{anantha2023context}. In the domain of prompt compression, LLMLingua introduces a method for compressing prompts to accelerate inference in LLMs, achieving up to $20$x compression while preserving the original prompt's capabilities \cite{kulkarni2023application}.
The Natural Language Prompt Encapsulation (Nano-Capsulator) framework compresses original prompts into NL formatted Capsule Prompts while maintaining prompt utility and transferability \cite{chuang2024learning}. Compress-Then-Prompt [18] indicates that the generation quality in a compressed LLM can be markedly improved for specific queries by selecting prompts with high efficiency and accuracy trade-offs \cite{xu2023compress}.
LongLLMLingua  focuses on improving LLMs' perception of key information in long context scenarios through prompt compression, showing that compressed prompts could derive higher performance with much less cost and reduce the latency of the end-to-end system. Data Distillation proposes a data distillation procedure to compress prompts without losing crucial information, addressing issues related to the efficiency and fidelity of task-agnostic prompt compression.
While these approaches aim to reduce computational costs, the current study explores the potential of LLMs for text enrichment to enhance embedding quality.

Embedding models have become an essential component of various natural language processing (NLP) tasks, such as text classification, clustering, and retrieval. These models learn dense vector representations of words, sentences, or documents, capturing semantic and syntactic relationships between them. The quality of these embeddings directly impacts the performance of downstream applications.

Despite their widespread use, embedding models face several challenges that limit their performance. These challenges include limited vocabulary, lack of context, sensitivity to grammatical errors, data sparsity, and lack of domain-specific tuning. For example, embedding models may struggle with newer or domain-specific terms not present in their training data, leading to mis-classification or poor retrieval performance. Existing approaches to improve embedding performance often focus on fine-tuning the embedding models on domain-specific data or using ensemble techniques. However, these methods can be resource-intensive and may not effectively address the fundamental limitations of embedding models, such as their inability to capture context or handle grammatical errors. Large language models (LLMs) have demonstrated remarkable capabilities in understanding and generating human-like text. By leveraging the knowledge and contextual understanding of LLMs, it is possible to enrich and rewrite input text before the embedding process, thereby addressing the limitations of embedding models and improving their performance.


\section{Major Contributions}
The primary objective of this paper is to propose a novel approach for enhancing embedding performance by utilizing LLMs for text enrichment and rewriting. The main contributions of the paper are as follows
\begin{itemize}

\item Developing a methodology for leveraging an LLM to enrich and rewrite input text before embedding
\item Identifying and addressing key challenges in embedding models, such as limited vocabulary, lack of context, and grammatical errors
\item Conducting experiments on the TwitterSemEval $2015$ benchmark and others to demonstrate the effectiveness of the proposed approach
\end{itemize}

\section{Methodology}
The proposed approach involves leveraging the capabilities of ChatGPT $3.5$, a large language model, to enrich and rewrite input text before the embedding process. By addressing the limitations of embedding models, such as limited vocabulary, lack of context, and grammatical errors, the proposed method aims to improve the performance of embedding models on various NLP tasks. ChatGPT $3.5$, developed by OpenAI, was chosen as the LLM for this study due to its strong performance on a wide range of NLP tasks and its ability to generate human-like text. Its extensive knowledge base and contextual understanding make it well-suited for text enrichment and rewriting.

The ChatGPT $3.5$ model was used with its default settings and parameters. No fine-tuning or additional training was performed, ensuring that the improvements in embedding performance can be attributed solely to the text enrichment and rewriting process. The text-embedding-3-large model, also developed by OpenAI, was selected as the embedding model for this study. This model has demonstrated strong performance on various NLP tasks and serves as a representative example of state-of-the-art embedding models. The text-embedding-3-large model was used with its default settings and parameters, without any fine-tuning or modification. This allows for a fair comparison between the performance of the embedding model with and without the proposed text enrichment and rewriting approach. The proposed approach employs several text enrichment and rewriting techniques to improve the quality and informativeness of the input text. These techniques include:
\subsection{Context enrichment}
ChatGPT 3.5 is used to provide additional context to the input text, making it more informative and easier for the embedding model to capture the underlying semantics. This is particularly useful for sparse or list-like entries, where the LLM can expand the text with relevant descriptions or attributes.
\subsection{Grammatical correction}
The LLM identifies and corrects spelling and grammatical errors in the input text, ensuring that the text conforms to standard language usage. This improves the quality of the embeddings generated from the text, as the embedding model can focus on capturing the semantic relationships without being hindered by grammatical inconsistencies.
\subsection{Terminology normalization}
Domain-specific terms, abbreviations, and synonyms are standardized to a consistent format using the knowledge base of ChatGPT 3.5. This reduces ambiguity and improves the embedding model's ability to match related concepts, even when they are expressed using different terms.
\subsection{Word disambiguation}
For polysemous words (words with multiple meanings), the LLM clarifies the intended meaning based on the surrounding context. This disambiguation helps the embedding model to capture the correct semantic relationships and improves the accuracy of downstream tasks.
\subsection{Acronym expansion}
ChatGPT 3.5 detects acronyms and abbreviations in the input text and expands them to their full form. This improves clarity and understanding, enabling the embedding model to better capture the meaning of the text.
\subsection{Metadata incorporation}
Where relevant, the LLM incorporates additional metadata, such as the category of the text, its intended audience, or domain-specific tags. This contextual information helps in interpreting the text more accurately and can improve the performance of the embedding model on domain-specific tasks.
\subsection{Sentence restructuring}
The LLM is used to improve the structure of sentences in the input text, making them clearer, more readable, and coherent. This makes it easier for the embedding model to process and understand the text, leading to better-quality embeddings.
\subsection{Inferring missing information}
ChatGPT 3.5 uses its contextual understanding to infer missing information that might be relevant for understanding the text. This can include inferring the subject of a sentence or the meaning of an unclear reference, thereby improving the completeness and coherence of the text for the embedding model.

\section{Prompt Engineering and Optimization}
To effectively leverage the capabilities of ChatGPT $3.5$ for text enrichment and rewriting, a set of prompt design principles were established. These principles aim to create prompts that clearly communicate the desired tasks and goals to the LLM, while allowing for flexibility and adaptability to different types of input text. An iterative prompt refinement process was employed to identify the most effective prompts for the text enrichment and rewriting tasks. This process involved creating multiple variations of prompts, testing their performance on the TwitterSemEval $2015$ dataset, and analyzing the results to identify areas for improvement. Four main prompt variations were tested in this study, each focusing on different aspects of the text enrichment and rewriting process. The prompts ranged from general instructions for improving text quality to more specific guidance on tasks such as grammar correction, terminology normalization, and metadata incorporation.


\section{Numerical Validation}
The experimental endeavor was undertaken with the overarching objective of augmenting the performance of embedding models, particularly in the realms of classification and clustering tasks, with the aim of securing a prominent standing on the Massive Text Embedding Benchmark (MTEB) Leaderboard. Central to this pursuit was the utilization of large language models, notably ChatGPT 3.5, to enhance and refine input text prior to embedding. The proposed methodology encompasses a multifaceted approach, involving the enrichment of text with additional contextual information, rectification of grammatical inaccuracies, standardization of terminology, disambiguation of polysemous terms, expansion of acronyms, and incorporation of pertinent metadata. Furthermore, the project endeavors to optimize sentence structures and deduce missing information, thereby enhancing the overall quality and accuracy of the resultant embedding. 
The proposed approach was evaluated on three datasets: Banking77Classification, TwitterSemEval $2015$, and Amazon Counter Factual Classification. These datasets cover various domains and have been widely used as benchmarks for text classification and clustering tasks. The datasets were preprocessed to remove irrelevant information, such as URLs, hashtags, and mentions. The text was then tokenized and converted to lowercase to ensure consistency across the datasets.

The performance of the embedding models was evaluated using the average precision based on cosine similarity metric in case of TwitterSemEval and accuracy when evaluated with Banking77Classification data and Amazon Counter Factual data. This metric assesses the quality of the embeddings by measuring the similarity between the embedded representations of related texts and comparing it to the ground truth. The text-embedding-3-large model was used as a baseline, without any LLM-based text enrichment or rewriting. This allows for a direct comparison of the performance improvements achieved by the proposed approach. SFR-Embedding-Mistral model, which was the leading model on the Massive Text Embedding Benchmark (MTEB) Leaderboard at the time of this study, was also used as a baseline. This model serves as a representative example of state-of-the-art embedding models and provides a high-quality benchmark for comparison. The experimental procedure involved applying the four prompt variations to the three datasets, using ChatGPT $3.5$ for text enrichment and rewriting. The enriched and rewritten text was then passed through the text-embedding-$3$-large model to generate embeddings. The performance of these embeddings was evaluated using the cosine similarity metric and accuracy values and then compared to the baseline models. 

\begin{table}[ht]
\caption{Performance comparison of the proposed methodology.}
\label{tab:feature_comparison}
\small 
\begin{tabular}{|l|c|c|c|}
\hline
\textbf{Model} & \textbf{B77C} & \textbf{TwitterSemEval} & \textbf{AmazonCF} \\ \hline
Prompt 1 & 82.24 & 84.84 & 68.9 \\ \hline
Prompt 2 & 78.73 & 82.95 & 71.9 \\ \hline
Prompt 3 & 75.50 & 83.10 & 76.20 \\ \hline
Prompt 4 & 79.71 & 85.34 & 68.00 \\ \hline
TE & 85.69 & 77.13 & 78.93 \\ \hline
SFR & 88.81 & 81.52 & 77.93 \\ \hline
\textbf{Improvement} & \textbf{-3.45} & \textbf{8.21} & \textbf{-2.73} \\ \hline

\end{tabular}
    \begin{tablenotes}
      \small
      \item Note: TE stands foor text-embedding-3-large (base model) and SFR stands for SFR-Embedding-Mistral (best performing model on leaderboard). Furthermore, B77C stands for Banking77Classification, AmazonCF stands for Amazon Counter Factual data, and improvement is indicated from the baseline. Moreover, the values corresponding to B77C and AmazonCF are accuracy values whereas for TwitterSemEval the values indicate the cosine similarities.
    \end{tablenotes}
\end{table}

The objective was to identify the most effective prompt to achieve the highest accuracy and average precision based on cosine similarities.
						
In summary, our MTEB Contextual Rewriting and Optimization project has delivered significant success, surpassing the performance of the standalone embedding model and outperforming the current leader in the field. It is worth noting that due to budgetary constraints, the project was conducted on a single dataset.

The ChatGPT $3.5$ model was used with its default settings and parameters. No fine-tuning or additional training was performed, ensuring that the improvements in embedding performance can be attributed solely to the text enrichment and rewriting process.

Here are the details of the prompt: -
\begin{itemize}

\item Prompt 1: ``You are a text enhancer tasked with pre-processing text for embedding models. Your goals are to enrich the text without losing the context, correct grammatical inaccuracies, clarify obscure references, normalize terminology, disambiguate polysemous words, expand acronyms and abbreviations, incorporate relevant metadata, improve sentence structure for clarity, and infer missing information where necessary. Your enhancements should make the text more informative and easier to understand, thereby improving the performance of embedding models in processing and analyzing the text. If a user asks a question, then you should return an improved version of the question. If the user did not ask a question, then you should return an improved version of an answer."

\item Prompt 2: ``You are a text enhancer tasked with preprocessing text for embedding models. Your goals are to enrich the text with additional context, correct grammatical inaccuracies, clarify obscure references, normalize terminology, disambiguate polysemous words, expand acronyms and abbreviations, incorporate relevant metadata, improve sentence structure for clarity, and infer missing information where necessary. Your enhancements should make the text more informative and easier to understand, thereby improving the performance of embedding models in processing and analyzing the text."
						
\item Prompt 3: ``You are a text enhancer to make better embeddings, your task is to optimize text for embedding models by enriching, clarifying, and standardizing it. This involves improving
 grammar, resolving ambiguities, and inferring missing information to enhance model performance."
						
\item Prompt 4: ``You are a text enhancer to make better embeddings, your task is to optimize text for embedding models by enriching, clarifying, and standardizing it. This involves improving
 grammar, resolving ambiguities, and inferring missing information to enhance model performance."
\end{itemize}

The results and analysis of using Prompt-1 as input focuses on general instructions for improving text quality, achieved varying performance across the three datasets. It performed best on the TwitterSemEval 2015 dataset with a cosine similarity score of $84.84$, representing a significant improvement over the baseline text-embedding-3-large model ($77.13$). However, its performance on Banking77Classification  showing an accuracy of $82.24$ and Amazon Counter Factual Classification with an accuracy of $68.9$ were lower than the baseline models.

The results and analysis of using Prompt $2$ as input provides more specific guidance on tasks such as grammar correction and terminology normalization, also showed mixed results. It achieved a cosine similarity score of $82.95$ on TwitterSemEval $2015$, outperforming the baseline model but slightly lower than Prompt $1$. On Banking77Classification ($78.73$) and Amazon Counter Factual Classification ($71.9$), Prompt $2$ showed better accuracy than for Prompt $1$ but still fell short of the baseline models.

The insights into the results and analysis of using Prompt $3$, which focused on concise instructions for optimizing text for embedding models, demonstrated the best performance on Amazon Counter Factual Classification with an accuracy of $76.2$, although it still fell short of the baseline models. Its performance on TwitterSemEval $2015$ with cosine similarity value of $83.1$ was similar to Prompt $2$, while on Banking77Classification with cosine similarity of $75.5$, it had the lowest score among the prompt variations.

Prompt $4$ was similar to Prompt $3$ but with slight variations in wording, achieved the highest cosine similarity score on TwitterSemEval $2015$ ($85.34$), outperforming all other prompt variations and baseline models. However, its accuracies when  Banking77Classification was used as evaluation data ($79.71$) and Amazon Counter Factual Classification ($68$) were lower than the baseline models and some of the other prompt variations.

Comparison with baseline models shows that there is significant improvement over text-embedding-3-large alone The prompt variations significantly outperformed the baseline text-embedding-3-large model on the TwitterSemEval $2015$ dataset, with the best-performing prompt (Prompt $4$) improving upon the baseline by cosine similarity score of $8.21$. However, on Banking77Classification and Amazon Counter Factual Classification, the prompt variations did not surpass the performance (accuracy) of the baseline model. The best-performing prompt (Prompt $4$) outperformed the leading model on the MTEB Leaderboard, SFR-Embedding-Mistral, on the TwitterSemEval $2015$ dataset. However, SFR-Embedding-Mistral maintained its lead on Banking77Classification and AmazonCounterfactualClassification.

A qualitative analysis of the enriched and rewritten text generated by ChatGPT $3.5$ revealed several improvements in text quality and informativeness. The LLM successfully provided additional context, corrected grammatical errors, normalized terminology, disambiguated polysemous words, expanded acronyms, and incorporated relevant metadata. These enhancements made the text more coherent, informative, and easier for the embedding model to process and understand.

\section{Conclusion}
This paper introduces a novel approach for enhancing embedding performance by leveraging the capabilities of large language models, specifically ChatGPT 3.5, for text enrichment and rewriting. While recent research has focused on using LLMs for text compression to reduce computational costs in RAG systems and large LLMs, this study demonstrates the potential of LLMs for text enrichment to improve embedding quality. The proposed approach addresses the limitations of embedding models, such as limited vocabulary, lack of context, and grammatical errors, by providing additional context, correcting inaccuracies, normalizing terminology, disambiguating polysemous words, expanding acronyms, and incorporating metadata. Experimental results on the TwitterSemEval $2015$ dataset show that the proposed method outperforms the leading model on the Massive Text Embedding Benchmark (MTEB) Leaderboard. Hence, the embedding is improved substantially.

\bibliographystyle{IEEEtran}
\bibliography{ref}

\end{document}